\documentclass[runningheads]{llncs}

\usepackage[utf8]{inputenc}
\usepackage{tikz}
\usepackage{lipsum}
\usepackage{graphicx}
\usepackage[linesnumbered,ruled,noend]{algorithm2e}
\SetKwInput{KwData}{Input}
\SetKwInput{KwResult}{Output}
\usepackage{pgfplots}
\pgfplotsset{compat=1.14}
\usetikzlibrary{arrows,shapes,snakes,automata,backgrounds,petri}

\begin{document}
\title{The Regularization of Small Sub-Constraint Satisfaction Problems}
\titlerunning{The Regularization of sub-CSPs}

\author{Sven L{\"o}ffler \and Ke Liu \and Petra Hofstedt}

\authorrunning{S. L{\"o}ffler et al.}

\institute{Brandenburg University of Technology Cottbus-Senftenberg, Germany\\
Department of Mathematics and Computer Science, MINT\\
Programming Languages and Compiler Construction Group\\
Konrad-Wachsmann-Allee 5, 03044 Cottbus \\
\email{\{sven.loeffler, liuke, hofstedt\}@b-tu.de}}

% check if algorithm lines are correct, I removed the ends 

\maketitle              % typeset the header of the contribution
\begin{abstract}
%This paper presents a new approach to optimize constraint satisfaction problems (CSPs). This is achieved by a greedy algorithm which detects sub-CSPs with a potential large number of conflicts and substitutes each of these sub-CSPs with only one locally consistent regular membership constraint.

%The purpose of this approach is to reduce the number of fails in the search tree in the solving process of a CSP. The consequent is an improving of the resolution speed of the original CSP and possibly a basic for work load balancing in the parallelization.

%This happens in a pre-processing process so that other optimizations like the use of redundancy or parallel computing operate without conflicts.

This paper describes a new approach on optimization of constraint satisfaction problems (CSPs) by means of substituting sub-CSPs with locally consistent regular membership constraints. The purpose of this approach is to reduce the number of fails in the resolution process, to improve the inferences made during search by the constraint solver by strengthening constraint propagation, and to maintain the level of propagation while reducing the cost of propagating the constraints. 
Our experimental results show improvements in terms of the resolution speed compared to the original CSPs and a competitiveness to the recent tabulation approach \cite{TabulationCP,Tabulation}. %presented in \cite{Tabulation,TabulationCP}. 
Besides, our approach can be realized in a preprocessing step, and therefore wouldn't collide with redundancy constraints or parallel computing if implemented.

\keywords{Constraint Programming, CSP, Refinement, Optimizations, Regular Membership Constraint, Regular CSPs}
\end{abstract}

\section{Introduction}
\label{sec:intro}
A CSP can be often described in several ways, each of which might consist of different types and combinations of constraints, which leads to various statistical results of the resolution, including the execution time, the number of fails, the number of backtracks, the number of nodes etc. The reason for this is, that the combination of constraints and their propagators have a significant impact on the shape and the size of the search tree.
Therefore, the diversity of models and constraints for a given CSP offers us an opportunity to improve the resolution process by using another model in which fewer fails occur during the resolution process.  
Based on this idea, previous works show that the performance of a constraint problem often can be improved by converting a sub-problem into a single constraint \cite{TabulationCP,Tabulation,Loeff17A,Loeff18A,IcaartReg}.

In this paper, we propose an algorithm which substitutes parts of CSPs by singleton, locally consistent constraints. 
In contrast to \cite{Tabulation}, the replacement is based on the regular membership constraint instead of the table constraint. 
Since our algorithm can be applied at the pre-processing stage, other approaches which accelerate the resolution process such as redundant modeling \cite{redundancy96}, parallel search \cite{ParallelSearch}, or parallel consistency \cite{Hamadi02} can be used in combination with ours.

The rest of this paper is organized as follows. In Section \ref{sec:prelim}, we introduce the necessary notions for the approach. In Section 3, the substitution of small sub-CSPs with the regular membership constraint is explained. 
In Section 4, the benefit of our regularization approach is shown in two case studies based on the Solitaire Battleships  Problem \cite{csplib:prob014} and the Black Hole Problem \cite{csplib:prob081}. Furthermore, we compare our results with the tabulation approach presented in \cite{Tabulation}. 
Finally, Section \ref{sec:future} concludes and proposes research directions for the future.
\begin{remark}
In this paper we will use the notion of a "\textit{regular constraint}" synonym for "\textit{regular membership constraint}".
\end{remark}

\section{Preliminaries}
\label{sec:prelim}

In this section, we introduce necessary definitions and methods for our regularization approach. 
%We consider constraint satisfaction problems like they are defined in \cite{Loeffler}.
We consider CSPs which are defined in the following way:\\

\noindent\textit{CSP} \cite{RDech}
A constraint satisfaction problem (CSP) is defined as a 3-tuple $P =
(X,D,C)$ with $X = \{x_1, x_2, \ldots, x_n \}$ is a set of variables, $D = \{D_1,D_2,\ldots$, $D_n\}$ is a set of finite domains where $D_i$ is the domain of $x_i$ and $C = \{c_1, c_2, \ldots, c_m \}$ is a set of primitive or global constraints covering between one and all variables in $X$.\\

\noindent Additionally, we define a sub-CSP $P_{sub}$ as a part of a CSP $P = (X, D, C)$ which covers only a part of the constraints and their variables.\\

\noindent\textit{Sub-CSP} Let $P = (X, D, C)$ be a CSP. For $C' \subseteq C$ we define $P_{sub} = (X',D',C')$ such that $X' = \bigcup_{c \in C'} scope(c)$ with corresponding domains $D' = \{D_i \mid x_i \in X'\} \subseteq D$, where the $scope$ of a constraint $c$ is defined as the set of variables which are part of the constraint $c$ \cite{RDech}\\

\noindent After we defined CSPs and sub-CSPs, we need a measure for the size of such a CSP or sub-CSP.\\

\noindent\textit{size($P$)} We define the maximal size $size(P)$ of a CSP $P = (X, D, C)$ as the product of the cardinalities of the domains of the CSP $P$, see Eq. \ref{eq:sizep}.

\begin{equation}
\label{eq:sizep}
size(P) = \prod_{i = 1}^{|X|} |D_i|
\end{equation}

\noindent The regular constraint, its propagation \cite{Stretch,Pesant2001,Pesant04} and deterministic finite automatons (DFAs) \cite{Hop} provide the basis of our approach.
We briefly review the notion of a deterministic finite automaton (DFA) and of the regular constraint.\\

\noindent\textit{DFA} \cite{Hop} A deterministic finite automaton (DFA) is a quintuple  
$M = (Q$, $\Sigma$, $\delta$, $q_0$, $F)$, where 
$Q$ is a finite set of states,
$\Sigma$ is the finite input alphabet,
$\delta$ is a transformation function $Q \times \Sigma \rightarrow Q$, 
$q_0 \in Q $ is the initial state,
and $F \subseteq Q$ is the set of final or accepting
 states.
A word $w \in \Sigma^*\;$%(w_1,w_2,\ldots, w_n)\end{math} 
is accepted by $M$,
i.e. $w\in L(M)$, if the corresponding DFA $M$ with the input $w$
stops in a final state $f \in F$.\\

\noindent\textit{Regular Constraint} \cite{Pesant04} Let $M = (Q$, $\Sigma$, $\delta$, $q_0$, $F)$ be a DFA and let $X = \{x_1, x_2, ..., x_n\}$ be a set of variables with $D(x_i) \subseteq \Sigma$ for $1 \leq i \leq n$. Then 
\begin{equation}
regular(X,M) = \{(d_1, ..., d_n) |  \forall i \; d_i \in D_i, d_1\circ d_2 \circ...\circ d_n \in L(M)\},
\end{equation}
i.e. every sequence $d_1...d_n$ of values for $x_1, ..., x_n$ must be a word of the regular language recognized by the DFA $M$, where $\circ$ is the concatenation of two words.

\section{Substitution of Constraints by Regular Constraints}
\label{sec: substitution}	

%In \cite{Loeffler} is presented that theoretical 
Previous work \cite{Loeff18A} has shown that each CSP can be transformed into an equivalent one with only one regular
constraint (rCSP), theoretically.   
In this section, we present a practical algorithm to transform the constraints of a sub-CSP $P_{sub}$ of a given CSP $P$ into a
regular constraint. For the reason of effectiveness the sub-CSP $P_{sub}$ should be much smaller than the original CSP
$P$ (size($P_{sub}) \ll size(P)$).

It is the aim to detect and substitute such sub-CSPs, which are preferably as big as possible but can be represented by a DFA which is as small as possible at the same time. An algorithm to detect such sub-CSPs must be developed in the future. 
Currently, we use the heuristics to find sub-CSPs given in \cite{TabulationCP}. Alternatively, an algorithm like Gottlobs
hypertree decomposition \cite{got08} or Ke Lius det-k-CP \cite{LiuLH17} can be used. 
%Alternative only one constraint or some constraints and all involved variables can be chosen as sub-CSP.

Our transformation algorithm obtains a sub-CSP $P_{sub} = (X', D', C')$ from CSP $P = (X$, $D$, $C)$ as input, where
$C' \subset C$, 
$X' = \{x_1,\ldots,x_n\} = \bigcup_{c \in C'}$ $scope(c)$ $\subset X$, $|X'| \ll |X|$ and 
$D' = \{D_1,\ldots,D_n\} \subset D$, where $D_i$ is the domain of variable $x_i, \forall i \in
\{1,2,\ldots,n\}$, and returns
a regular constraint which is equivalent to the constraints in $C'$. 
Our regularization algorithm has two phases:
\begin{enumerate}
\item Solve the detected/given sub CSP $P_{sub}$.
\item Transform all solutions $S = \{s_1,s_2,\ldots,s_k\}$ of the sub-CSP $P_{sub}$  into a regular constraint.
\end{enumerate}
The first phase is obvious. Notice that the sub-CSP $P_{sub}$ should be much
smaller than the original CSP $P$, otherwise the solving step would be too time consuming. 

We continue with a description of the second phase.
Let $S = \{s_1,s_2,\ldots,s_k\}$ be the
set of all solutions of $P_{sub}$ calculated in step one. %
Every solution $s_j$, $j \in \{1,2,\ldots,k\}$ consists of $n$ values
$s_{i,j}$, $i \in \{1,2,\ldots,n\}$, cf. Table \ref{tab:Solution}.
\begin{table}
\caption{\label{tab:Solution} The solutions $s_1, ..., s_k$ of the sub-CSP $P_{sub}$}
\centering
\begin{tabular}{l|l l l l}
\hline
\ $S$ & \ $s_1$ & \ $s_2$ & \ \ldots & \ $s_k$\\
\hline
\ $x_1$ \ & \ $s_{1,1}$ \ & \ $s_{1,2}$ \ & \ \ldots \ & \ $s_{1,k}$ \ \\
\ $x_2$ & \ $s_{2,1}$ & \ $s_{2,2}$ & \ \ldots & \ $s_{2,k}$\\
\ \ \vdots & \ \ \vdots  & \ \ \vdots  & \ $\ddots$ & \ \ \vdots \\
\ $x_n$ & \ $s_{n,1}$ & \ $s_{n,2}$ & \ \ldots & \ $s_{n,k}$\\
\hline
\end{tabular}
\end{table}

\noindent To define a deterministic finite automaton as the basis for the regular constraint, we need the set $T = \{T_1,\ldots,T_n\}$ of prefix sets of
all solutions of $P_{sub}$, where all elements in $T_i$ are concatenations of the $i$ first values of a solution $s \in S$ (see Eq. \ref{eq1}):   
\begin{equation}
\label{eq1}
%T_i = \bigcup\limits_{l = 1}^k getFirstI(s_l,i)\ |\ i \in \{1,2,\ldots,n-1\} 
T_i = \bigcup\limits_{l=1}^k \{s_{1,l} \ \circ \ s_{2,l} \ \circ \  \ldots \ \circ \
s_{i,l} \; | \; \forall i \in \{1,\ldots,n\}\}
\end{equation}
This results in e.g. $T_1=\{s_{1,1}, s_{1,2},$ $\ldots, s_{1,k}\}$, $T_2 = \{s_{1,1} \circ s_{2,1}, s_{1,2} \circ s_{2,2},
\ldots, s_{1,k} \circ s_{2,k}\}$, $T_n = S$.
Note that we enumerate the elements in each $T_i$ from $1$ to $k$ but actually they mostly have fewer elements then $k$ for the reason that multiple occurrences of elements do not occur in sets. It follows $|T_1| \leq |T_2| \leq \ldots \leq |T_n| = k$.  

% The method $getFirstI(s_l, i)$ returns the first $i$ values of the
% solution $s_l$. The first set $T_0$ is set to $\{\{\emptyset\}\}$.

\medskip

% For each element $t_{i,l},\ l \in \{1,2,\ldots,|T_i|\}$ of each set $T_i,\ i \in \{1,\ldots,n-1\}$ a state $q_{i,l}$ for the DFA is created, which represents the solution prefix $t_{i,l}$. Furthermore, the initial state $q_{start}$ and the final state $q_{end}$ are added to $Q$. Thus, the \textit{set of states $Q$ of the DFA} is

For each element $t$ of each set $T_i,\ i \in \{1,\ldots,n-1\}$ a state $q_{t}$ for the DFA is created, which represents the solution prefix $t$. Furthermore, the initial state $q_{start}$ and the final state $q_{end}$ (representing all solutions $S = T_n$ of $P_{sub}$) are added to $Q$. Thus, the \textit{set of states $Q$ of the DFA} is
%
%$$Q = \{q_{i,l}\ |\ \forall i \in \{1,\ldots,n-1\}, \forall l \in \{1,2,\ldots,|T_i|\}\} \cup \{q_{start},q_{end}\}.$$
$$Q = \{q_{t}\ |\ t \in T_i, i \in \{1,2,\ldots,n-1\}\} \cup \{q_{start},q_{end}\}.$$
The \textit{initial state} is $q_{start}$ and $F= \{q_{end}\}$ is the \textit{set of final states}.

\medskip

\textit{The alphabet $\Sigma$ of the DFA} is the union of all
domains of the variables of $X'$:
\begin{equation}
\Sigma = \bigcup\limits_{D_i \in D'} D_i
\end{equation}

\medskip

Finally, we define the \textit{transition function $\delta$} as follows:

\begin{itemize}

\item 

Let %
%\begin{minipage}[t]{0.8\linewidth}
%\begin{tabbing}
$t \in T_{1}$.
%\end{tabbing}
%\end{minipage}
Then it holds
\begin{equation}
\label{eq9}
\delta(q_{start},t) = q_t
\end{equation}

\item
Let %
%\begin{minipage}[t]{0.8\linewidth}
%\begin{tabbing}
% $t_{i-1,l_1} = a_{1,l_1} a_{2,l_1} \ldots a_{i-1,l_1}\ \ \ \ \ \in T_{i-1}$\\
% $t_{i,l_2}\ \ \ = a_{1,l_1} a_{2,l_1} \ldots a_{i-1,l_1} b_{i,l_2} \in T_{i}$\\
% $i\  \in \{2,\ldots, n-1\}$\\
% $l_1 \in \{1,\ldots,|T_{i-1}|\}$\\
% $l_2 \in \{1,\ldots,|T_{i}|\}$
$t_{i-1}$ be an element in $T_{i-1}$, 
$t_{i}$ be an element in $T_{i}$,  
$i \in \{2,\ldots, n-1\}$ and
$w \in D_i$ with $t_{i} = t_{i-1} \circ w$.
%\end{tabbing}
%\end{minipage}
%
%\smallskip
%
Then it holds 
\begin{equation}
\label{eq10}
% \delta(q_{i-1, l_1},b_{i,l_2})=q_{i,l_2}
\delta(q_{t_{i-1}},w)=q_{t_i}
\end{equation}

\item
Let %
%\begin{minipage}[t]{0.8\linewidth}
%\begin{tabbing}
$t_{n-1}$ be an element in $T_{n-1}$,
$t_{n}$ be an element in $T_{n} = S$ and
$w \in D_n$ with $t_{n} = t_{n-1} \circ w$.
%\end{tabbing}
%\end{minipage}
%
%\smallskip
%
Then it holds 
\begin{equation}
\label{eq11}
\delta(q_{t_{n-1}},w)= q_{end}
\end{equation}
\end{itemize}

%\begin{equation}
%\begin{tabular}{p{3cm} p{8cm}}
%$(q_{i,a}, d_{j}) \rightarrow q_{i+1,b}$ 
%% & if $q_{i,a}$ exists in $P_i$ and \newline $q_{i+1,b}$
%% exists in $P_{i+1}$ and \newline $P_{i,a}d_i = P_{i+1,b}$ 
%& $\forall i \in
%\{0,1,\ldots,n-2\}$, $\forall d_{j} \in D_{i+1}$, \newline $\forall a \in \{1,2,\ldots,|T_i|\}$
% $\forall b \in \{1,2,\ldots,|T_{i+1}|\}$ \newline\hspace*{0.5cm}where $t_{i,a}d_{j} = t_{i+1,b}$
%  \end{tabular}
%\end{equation}
%\begin{equation}
%	\begin{tabular}{p{3cm} p{8cm}}
%$(q_{n-1,a}, d_j) \rightarrow q_{end}$ 
%% & if $q_{i,a}$ exists in $P_i$ and \newline $q_{i+1,b}$
%% exists in $P_{i+1}$ and \newline $P_{i,a}d_i = P_{i+1,b}$ 
%& $\forall d_{j} \in D_{n}$, \newline $\forall a \in \{1,2,\ldots,|T_{n-1}|\}$
% \newline\hspace*{0.5cm}where $t_{i,a}d_{j} \in S$
%  \end{tabular}
%  \end{equation}

\medskip

\noindent This altogether provides the DFA %
$M = (Q, \Sigma, \delta, q_{start},\{q_{end}\})$. 
The constraint $regular(X', M)$ can be used as a replacement for the
constraints of $C'$ in the original CSP $P$.
\begin{remark} This algorithm is only useful for sub-CSPs $P_{sub}$ which are
proper subsets of the original CSP $P$ ($size(P_{sub}) \ll size(P)$). Solving a sub-problem $P_{sub}$ and
finding all solutions is also an NP-hard problem. 
Nevertheless, due to the exponential growth of constraint problems, sub-problems with smaller size than the original problem can be solved significantly faster.
\end{remark}
%In our greedy algorithm for the regularization of CSPs (see \ref{a greedy algo}) we can use these
%education regulations in step three. Because the goal of the algorithm is to reduce the number of
%fails in the search tree we only substitute the constraints of the sub-CSPs if the sub-CSP has some
%fails (especially bigger than the entered value for $minFailNumber$), otherwise we don't change the
%constraints in the original CSP $P$.  

\section{Examples and Experimental results}
\label{sect:example}

After presenting our approach to transform the constraints of small sub-CPSs into a regular constraint, we want to show two case studies to underline its benefits. For this, we use the Black Hole Problem \cite{csplib:prob081} and the Solitaire Battleships Problem \cite{csplib:prob014} from the CSPlib.

All the experiments are set up on a DELL laptop with an Intel i7-4610M CPU, 3.00GHz, with 16 GB 1600 MHz DDR3 and running under Windows 7 professional with service pack 1. The algorithms are implemented in Java under JDK version 1.8.0\_191 and Choco Solver \cite{choco}. 
We used the $DowOverWDeg$ search strategy which is explained in \cite{BoussemartHLS04} and is used as default search strategy in the Choco Solver \cite{choco}.

\subsection{The Black Hole Problem}
Black Hole is a common card game, where all 52 cards are played one after 
the other from seventeen face-up fans of three cards into a discard pile named ‘black hole’, which contains at the beginning only the card $A\spadesuit$. All cards are visible at all times. A card can be played into the ‘black hole’ if it is adjacent
in rank to the previous card (colors are not important). The goal is to play all cards into the Black Hole. 

Black Hole was modelled for a variety of solvers
by Gent et al. \cite{DBLP:Gent}. We use the
simplest and most declarative model of Dekker et al. \cite{DBLP:Dekker}, where two variables $a$ and
$b$ represent adjacent cards if $|a-b| \; mod \; 13 \in \{1,12\}$.

The heuristic \textit{Weak Propagation}, presented in \cite{TabulationCP}, detects the adjacency constraints as replaceable\footnote{In \cite{TabulationCP}, the detected constraints are substituted by table constraints, in contrast to the here presented approach; we will substitute them with regular constraints.}. 
For our benchmark suite we computed 50 different instances of the Black Hole Problem, where 49 instances are randomly created (so the position of every card in the 17 fans is random) and the remaining instance has an enumerated card distribution ($1\spadesuit$, $2\spadesuit$, ..., $K\spadesuit$, $A\clubsuit$, $1\clubsuit$,..., $K\clubsuit$, $A\heartsuit$, $1\heartsuit$, ..., $K\heartsuit$, $A\diamondsuit$, $1\diamondsuit$, ..., $K\diamondsuit$). 

For all instances, we limited the solution time to 10 minutes and each problem was solved in 4 ways: 
\begin{enumerate}
\item \textit{Original}: The problem was modelled as described in \cite{DBLP:Dekker}.
\item \textit{Table}: The detected adjacency constraints were substituted by table constraints.
\item \textit{Regular}: The detected adjacency constraints were substituted by regular constraints.
\item \textit{RegularIntersected}: The detected adjacency constraints were substituted by only one regular constraint. The single regular constraint was created by the intersection of the underlying automatons of the substituted regular constraints from item (3) \textit{Regular} as given above.
\end{enumerate}

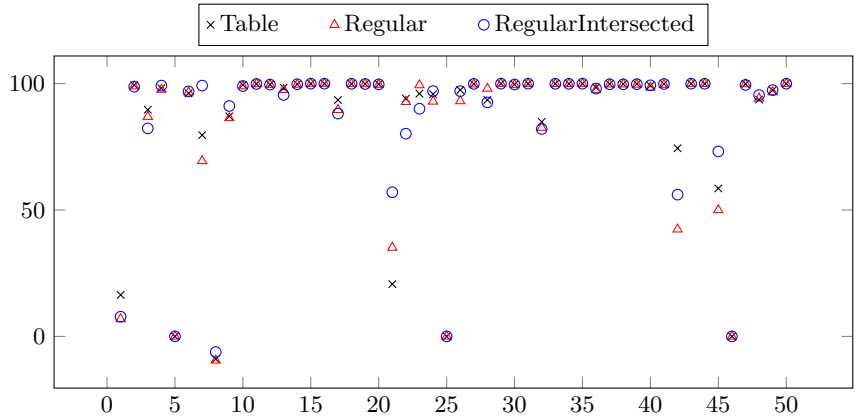
\begin{figure}
\begin{tikzpicture}
\pgfplotsset{every axis legend/.append style={at={(0.5,1.03)},anchor=south}}
\begin{axis}[width= \textwidth,height=6cm, legend columns=4]
\addplot[mark=x,only marks] table [only marks] 
{
x    y    
1	16.416
2	99.389
3	89.677
4	98.175
5	0.000
6	96.174
7	79.630
8	-9.000
9	87.075
10	99.514
11	99.947
12	99.638
13	98.391
14	99.760
15	99.988
16	99.984
17	93.536
18	99.942
19	99.899
20	99.815
21	20.659
22	94.053
23	96.044
24	95.829
25	0.000
26	97.466
27	99.900
28	93.546
29	99.985
30	99.686
31	99.979
32	84.861
33	99.947
34	99.964
35	99.984
36	98.486
37	99.880
38	99.857
39	99.909
40	99.235
41	99.865
42	74.427
43	99.987
44	99.962
45	58.537
46	0.000
47	99.884
48	93.675
49	97.359
50	99.984
};
\addplot[mark=triangle, red,only marks] table [only marks] 
{
x    y 
1	6.868
2	98.956
3	86.897
4	97.634
5	0.000
6	96.599
7	69.406
8	-9.500
9	86.395   
10	99.000
11	99.885
12	99.565
13	97.542
14	99.626
15	99.987
16	99.971
17	89.633
18	99.889
19	99.818
20	99.604
21	35.125
22	92.794
23	99.419
24	92.939
25	0.000
26	93.094
27	99.723
28	98.061
29	99.986
30	99.449
31	99.977
32	82.593
33	99.889
34	99.908
35	99.984
36	97.566
37	99.751
38	99.744
39	99.646
40	98.498
41	99.580
42	42.354
43	99.980
44	99.931
45	50.000
46	0.000
47	99.806
48	94.153
49	96.646
50	99.985  
};
\addplot[mark=o, blue,only marks] table [only marks] 
{
x    y 
1	7.784
2	98.761
3	82.283
4	99.217
5	0.000
6	96.918
7	99.218
8	-6.200
9	91.156
10	99.036
11	99.862
12	99.613
13	95.519
14	99.747
15	99.986
16	99.984
17	88.155
18	99.962
19	99.863
20	99.711
21	57.027
22	80.192
23	90.053
24	97.046
25	0.000
26	97.024
27	99.904
28	92.586
29	99.986
30	99.653
31	99.981
32	82.043
33	99.954
34	99.933
35	99.986
36	98.022
37	99.750
38	99.726
39	99.782
40	99.332
41	99.861
42	56.070
43	99.987
44	99.948
45	73.171
46	0.000
47	99.476
48	95.465
49	97.431
50	99.985
};
\legend{Table \hspace*{0.5cm}, Regular \hspace*{0.5cm}, RegularIntersected}
\end{axis}
\end{tikzpicture}
\caption{The time improvements (in \%) of the \textit{Table}, \textit{Regular} and \textit{RegularIntersected} models for finding the first solution of each instance of the Black Hole Problem in comparison to the \textit{Original} model (0\%).} \label{fig1}
\end{figure}

\begin{table}
\caption{Overwiev of the Black Hole bechmark.}\label{tab1}
\begin{tabular}{|l|r|r|r|r|}
\hline
&Ave. Solution Time & Ave. Imp. in \% & $\#$ Fastest  & $\#$ Sol. Instances
\\
\hline
\textit{Original} &	516.432s	&- &1	&7\\\hline
\textit{Table} &	58.796s & 83.413\% &	25&	47\\\hline
\textit{Regular}	& 63.679s& 82.054\%&	2&	47\\\hline
\textit{RegularIntersected} &	54.883s& 84.165\%	&19&	47\\\hline
\end{tabular}
\end{table}

\noindent Figure \ref{fig1} shows the time improvements (in \%) of the three substituted models (\textit{Table}, \textit{Regular} and \textit{RegularIntersected}) in comparison to the \textit{Original} model when the first solution is searched. In 49 of 50 cases all modified models are better than the original. The only exception is sample case 8, where the original approach is $62 - 95\%$ faster than the substituted ones\footnote{For case 8 exists a deterioration of 65\% (90\%, 95\%) for the \textit{RegularIntersected} (\textit{Table} and \textit{Regular}) approach. To keep the graphic small the negative values were drawn in $\frac{1}{10}$ of the real distance. In cases 5, 25 and 46 none of the four models found a solution in the time bounds of 10 minutes.}. Table \ref{tab1} shows that the \textit{Table} approach was 25 times, the \textit{RegularIntersected} approach was 19 times, the \textit{Regular} approach was two times and the \textit{Original} approach was one time the fastest. In average we could reach the first solution $83.413\%$, $82.054\%$ or $84.165\%$ faster than the \textit{Original} approach and we could solve many more problem instances with the substitution approaches in the time limit in comparison to the \textit{Original} model (47 instead of 7).

\subsection{The Solitaire Battleships Problem}
The Solitaire Battleships Problem is a famous symbol puzzle, where several ships with different sizes must be placed on a two-dimensional grid.
The ships may be oriented horizontally or vertically, and no two ships will occupy adjacent grid squares, not even diagonally. Numerical values along the right hand side of and below the grid indicate the number of grid squares in the corresponding rows and columns that are occupied by vessels (see more details in \cite{csplib:prob014}).

We created an equivalent Choco version of the \mbox{\textit{MiniZinc}} model given in \cite{csplib:prob014} and tested the introductory example and the 35 instances given in the "sb\_Mini-Zinc\_Benchmarks.zip" from \cite{csplib:prob014}. We indicated the "spacing constraints", the "ship shape constraints" and the "count number of bigger ships constraints" as potential good candidates for a substitution by regular (or table) constraints. 

For all instances we limited the solution time to 30 minutes and each problem
was solved in five ways:

\begin{enumerate}
\item \textit{Original}: The problem was modelled as described in \cite{csplib:prob014}.
\item \textit{Table}: With reference to \cite{csplib:prob014}, the single lines 75 to 80 of the "spacing constraints", the single lines 86 to 89 and the three lines 91 to 93 together of the the "ship shape constraints" and each two lines 117 to 118 and 122 to 123 together of the "count number of bigger ships constraints" were each substituted by singleton table constraints.
%\item \textit{Table}: The lines of the "spacing constraints", the lines 86, 87, 88, 89 and 91 to 93 of the "ship shape constraints" and the lines 117 to 118 and 122 to 123 of the "count number of bigger ships constraints" in \cite{csplib:prob014} were each substituted by one table constraint. 
\item \textit{Regular}: The lines enumerated in \textit{Table} were substituted with regular constraints.
\item \textit{RegularIntersected}: Equivalently to \textit{Regular}, except the partial constraints in "count number of bigger ships constraints" which count the number of ships of size $s$ in a row, respectively in a column, were combined each to one regular constraint.
\item \textit{TableRegularIntersected}: There, we have the same combined regular constraints (for representing the "count number of bigger ships constraints") as described in \textit{RegularIntersected},
 but, apart from that, use the table constraints described in \textit{Table} (for representing the "spacing constraints" and "ship shape constraints").
\end{enumerate}

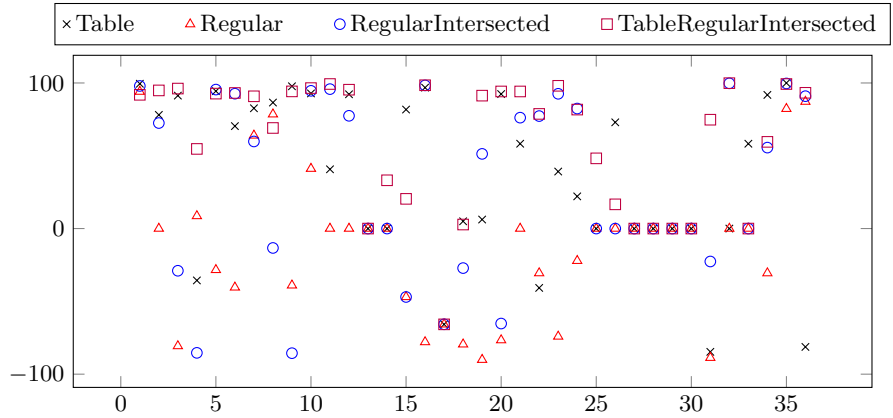
\begin{figure}[t!]
\begin{tikzpicture}
\pgfplotsset{every axis legend/.append style={at={(0.5,1.03)},anchor=south}}
\begin{axis}[width= \textwidth,height=6cm, legend columns=4]
\addplot[mark=x,only marks] table [only marks] 
{
1	99.471
2	78.078
3	91.323
4	-35.616
5	94.389
6	70.374
7	82.625
8	86.560
9	97.729
10	92.823
11	40.708
12	92.227
13	0.000
14	0.000
15	81.746
16	96.802
17	-65.803
18	4.993
19	6.200
20	92.382
21	58.280
22	-40.746
23	39.169
24	22.110
25	0.000
26	72.967
27	0.000
28	0.000
29	0.000
30	0.000
31	-84.676
32	0.000
33	58.273
34	91.797
35	99.995
36	-81.290
};

\addplot[mark=triangle, red,only marks] table [only marks] 
{
1	94.052
2	0.000
3	-80.692
4	8.625
5	-28.390
6	-40.407
7	63.960
8	78.505
9	-39.001
10	41.140
11	0.000
12	0.000
13	0.000
14	0.000
15	-46.973
16	-77.945
17	-65.760
18	-79.497
19	-90.087
20	-76.625
21	0.000
22	-30.596
23	-74.121
24	-22.142
25	0.000
26	0.000
27	0.000
28	0.000
29	0.000
30	0.000
31	-88.731
32	0.000
33	0.000
34	-30.616
35	82.267
36	87.190
};

\addplot[mark=o, blue,only marks] table [only marks] 
{
1	97.949
2	72.483
3	-28.975
4	-85.366
5	95.465
6	92.641
7	59.824
8	-13.321
9	-85.592
10	94.613
11	95.777
12	77.453
13	0.000
14	0.000
15	-46.985
16	98.330
17	-65.749
18	-27.147
19	51.296
20	-65.233
21	76.172
22	77.264
23	92.571
24	82.433
25	0.000
26	0.000
27	0.000
28	0.000
29	0.000
30	0.000
31	-22.616
32	99.751
33	0.000
34	55.606
35	99.666
36	90.934
};

\addplot[mark=square, purple,only marks] table [only marks] 
{
1	91.869
2	94.887
3	96.158
4	54.699
5	92.689
6	93.127
7	90.852
8	69.087
9	94.220
10	96.491
11	99.285
12	95.373
13	0.000
14	33.165
15	20.402
16	98.451
17	-65.790
18	2.773
19	91.293
20	94.170
21	94.170
22	78.741
23	98.056
24	81.724
25	48.226
26	16.638
27	0.000
28	0.000
29	0.000
30	0.000
31	74.765
32	99.870
33	0.000
34	59.546
35	99.282
36	93.248
};

\legend{Table \hspace*{0.5cm}, Regular \hspace*{0.5cm}, RegularIntersected \hspace*{0.5cm}, TableRegularIntersected}
\end{axis}
\end{tikzpicture}
\caption{The time improvements (in \%) of the \textit{Table}, \textit{Regular}, \textit{RegularIntersected} and \textit{TableRegularIntersected} models for finding the first solution of each instance of the Solitaire Battleships Problem in comparison to the \textit{Original} model (0\%).} \label{fig2}
\end{figure}

\begin{table}
\caption{Overwiev of the Black Hole bechmark.}\label{tab2}
\begin{tabular}{|l|r|r|r|r|}
\hline
&Ave. Sol. Time & Ave. Imp. in \% & $\#$ Fastest  & $\#$ Sol. Instances
\\
\hline
\textit{Original} &	935.168& -&	1&	23\\\hline
\textit{Table} &	632.955	& 37.303\%
 &9	&28\\\hline
\textit{Regular}	 &1120.421 & -11.551\%
	&0	&17\\\hline
\textit{RegularIntersected} &	677.923	& 29.701\%
& 2&	26\\\hline
\textit{TableRegularIntersected}  &	507.820	& 60.763\%
& 19&	30\\\hline
\end{tabular}
\end{table}

\noindent Figure \ref{fig2} shows that the results for the Solitaire Battleships Problem are not that clear as the results for the Black Hole Problem. A look into table \ref{tab2} reveals that the improvements for finding a first solution are very streaky. 
%The simple \textit{Regular} approach is on average a deterioration in this case. 
The \textit{Table} approach was the best approach, if using only one substitution style (tabulation or regularization). It found the first solution in 9 cases as fastest and was in average 37\% faster than the original approach. The \textit{Regular} approach slows the solution process down here but the \textit{RegularIntersected} approach leads again to a speed up (2 times fastest approach, 29.701\% better as the \textit{Original} approach), which is not much worse than the speed up from the \textit{Table} approach. 

The \textit{TableRegularIntersected} approach shows that a combination of regularization and tabulation can lead to a significant improvement. Here it was the best approach. It could solve the most problems (30), could find most often as fastest the first solution (19) and had in average the biggest time improvement (60.763\%). 

\remark{The \textit{TableRegularIntersected} approach was not calculated fully automatically here, but it shows the potential of both approaches in combination. Future work has to be done automate the combination of both approaches.}

\remark{In the evaluation, we did not present the needed time for the trans-
formations. Depending on the specific CSPs, we observed big differences
in the necessary transformation times. In our case, the total transformation time
needed for all transformations were in all Black Hole instances less than three and in all Solitaire Battleship instances less than four seconds.
Because the transformation time can be neglected in comparison to the solution
time (less than three respectively four seconds vs. 10 respectively 30 minutes) we did not figure out them explicitly.}

\section{Conclusion and Future Work}
\label{sec:future}

We presented a new approach for the optimization of general CSPs using the regular constraint.
For this a suitable sub-set of constraints are detected (for example with heuristics presented in \cite{TabulationCP}), solved separately and transformed into a regular constraint. Two benchmarks stress the benefit of this approach in comparison to the original problems and the competitiveness to the tabulation approach presented in \cite{Tabulation}. Furthermore, our benchmarks indicate the potential of a combination of both approaches.

In the future we will research heuristics, for finding sub-CSPs which are especially suitable for the regularization approach.
Besides, we want to consider the idea of direct transformations from several global constraints to equivalent regular constraints \cite{Loeff18A} and the combination of regular constraints transformed from global constraints with regular constraints transformed from sub-CSPs. We expect that this combination approach can be applied more often than the tabulation approach \cite{Tabulation}, because big sub-CSPs can be represented by a small DFA often; in contrast to this a table constraint always needs to store all solution tuples. 
Therefore, the regularization approach looks more promissing for big problems.
%This allows the use of regularization also for bigger sub-CSPs.

The most obvious next step is a detailed comparison of the regularization approach with the tabulation approach and the formulation of heuristics which suggest when which approach is more advantageous.


\bibliographystyle{splncs04}
\bibliography{bibliography}

\begin{thebibliography}{10}
\providecommand{\url}[1]{\texttt{#1}}
\providecommand{\urlprefix}{URL }
\providecommand{\doi}[1]{https://doi.org/#1}

\bibitem{TabulationCP}
Akg{\"{u}}n, {\"{O}}., Gent, I.P., Jefferson, C., Miguel, I., Nightingale, P.,
  Salamon, A.Z.: Automatic discovery and exploitation of promising subproblems
  for tabulation. In: Principles and Practice of Constraint Programming - 24th
  International Conference, {CP} 2018, Lille, France, August 27-31, 2018,
  Proceedings. pp. 3--12 (2018),
  \url{https://doi.org/10.1007/978-3-319-98334-9\_1}

\bibitem{BoussemartHLS04}
Boussemart, F., Hemery, F., Lecoutre, C., Sais, L.: Boosting systematic search
  by weighting constraints. In: de~M{\'{a}}ntaras, R.L., Saitta, L. (eds.)
  Proceedings of the 16th Eureopean Conference on Artificial Intelligence,
  ECAI'2004, including Prestigious Applicants of Intelligent Systems, {PAIS}
  2004, Valencia, Spain, August 22-27, 2004. pp. 146--150. {IOS} Press (2004)

\bibitem{redundancy96}
Cheng, B.M.W., Lee, J.H.M., Wu, J.C.K.: Speeding up constraint propagation by
  redundant modeling. In: Freuder, E.C. (ed.) Principles and Practice of
  Constraint Programming - CP96. pp. 91--103. Springer Berlin Heidelberg,
  Berlin, Heidelberg (1996)

\bibitem{RDech}
Dechter, R.: Constraint processing. Elsevier Morgan Kaufmann (2003)

\bibitem{DBLP:Dekker}
Dekker, J.J., Bj{\"{o}}rdal, G., Carlsson, M., Flener, P., Monette, J.:
  Auto-tabling for subproblem presolving in minizinc. Constraints
  \textbf{22}(4),  512--529 (2017),
  \url{https://doi.org/10.1007/s10601-017-9270-5}

\bibitem{csplib:prob014}
Gent, I.: {CSPLib} problem 014: Solitaire battleships.
  \url{http://www.csplib.org/Problems/prob014}, last visited on 2019-05-07

\bibitem{DBLP:Gent}
Gent, I.P., Jefferson, C., Kelsey, T., Lynce, I., Miguel, I., Nightingale, P.,
  Smith, B.M., Tarim, A.: Search in the patience game 'black hole'. {AI}
  Communications  \textbf{20}(3),  211--226 (2007),
  \url{http://content.iospress.com/articles/ai-communications/aic405}

\bibitem{got08}
Gottlob, G., Samer, M.: A backtracking-based algorithm for hypertree
  decomposition. Journal of Experimental Algorithmics (JEA)  \textbf{13} (2008)

\bibitem{Hamadi02}
Hamadi, Y.: Optimal distributed arc-consistency. Constraints  \textbf{7}(3-4),
  367--385 (2002), \url{https://doi.org/10.1023/A:1020594125144}

\bibitem{Stretch}
Hellsten, L., Pesant, G., van Beek, P.: A domain consistency algorithm for the
  stretch constraint. In: Wallace, M. (ed.) Principles and Practice of
  Constraint Programming - {CP} 2004. Lecture Notes in Computer Science,
  vol.~3258, pp. 290--304. Springer (2004)

\bibitem{Hop}
Hopcroft, J.E., Ullman, J.D.: Introduction to Automata Theory, Languages and
  Computation. Addison-Wesley (1979)

\bibitem{Tabulation}
Lecoutre, C.: {STR2:} optimized simple tabular reduction for table constraints.
  Constraints  \textbf{16}(4),  341--371 (2011),
  \url{https://doi.org/10.1007/s10601-011-9107-6}

\bibitem{LiuLH17}
Liu, K., L{\"{o}}ffler, S., Hofstedt, P.: Hypertree decomposition: The first
  step towards parallel constraint solving. In: Declarative Programming and
  Knowledge Management - Conference on Declarative Programming, {DECLARE} 2017,
  Unifying INAP, WFLP, and WLP, W{\"{u}}rzburg, Germany, September 19-22, 2017,
  Revised Selected Papers. pp. 81--94 (2017),
  \url{https://doi.org/10.1007/978-3-030-00801-7\_6}

\bibitem{Loeff17A}
L{\"{o}}ffler, S., Liu, K., Hofstedt, P.: The power of regular constraints in
  csps. In: 47. Jahrestagung der Gesellschaft f{\"{u}}r Informatik, Informatik
  2017, Chemnitz, Germany, September 25-29, 2017. pp. 603--614 (2017),
  \url{https://doi.org/10.18420/in2017\_57}

\bibitem{Loeff18A}
L{\"{o}}ffler, S., Liu, K., Hofstedt, P.: The regularization of csps for
  rostering, planning and resource management problems. In: Artificial
  Intelligence Applications and Innovations - 14th {IFIP} {WG} 12.5
  International Conference, {AIAI} 2018, Rhodes, Greece, May 25-27, 2018,
  Proceedings. pp. 209--218 (2018),
  \url{https://doi.org/10.1007/978-3-319-92007-8\_18}

\bibitem{IcaartReg}
L{\"{o}}ffler, S., Liu, K., Hofstedt, P.: A meta constraint satisfaction
  optimization problem for the optimization of regular constraint satisfaction
  problems. In: Rocha, A.P., Steels, L., van~den Herik, J. (eds.) Proceedings
  of the 11th International Conference on Agents and Artificial Intelligence,
  {ICAART} 2019, Volume 2, Prague, Czech Republic, February 19-21, 2019. pp.
  435--442. SciTePress (2019). \doi{10.5220/0007260204350442},
  \url{https://doi.org/10.5220/0007260204350442}

\bibitem{csplib:prob081}
Nightingale, P.: {CSPLib} problem 081: Black hole.
  \url{http://www.csplib.org/Problems/prob081}, last visited on 2019-05-07

\bibitem{Pesant2001}
Pesant, G.: A filtering algorithm for the stretch constraint. In: Walsh, T.
  (ed.) Principles and Practice of Constraint Programming - {CP} 2001. Lecture
  Notes in Computer Science, vol.~2239, pp. 183--195. Springer (2001)

\bibitem{Pesant04}
Pesant, G.: A regular language membership constraint for finite sequences of
  variables. In: Wallace, M. (ed.) Principles and Practice of Constraint
  Programming - {CP} 2004. Lecture Notes in Computer Science, vol.~3258, pp.
  482--495. Springer (2004)

\bibitem{choco}
Prud'homme, C., Fages, J.G., Lorca, X.: Choco Documentation. TASC, INRIA
  Rennes, LINA CNRS UMR 6241, COSLING S.A.S. (2016),
  \url{http://www.choco-solver.org/}, last visited 2019-05-07

\bibitem{ParallelSearch}
R{\'e}gin, J.C., Rezgui, M., Malapert, A.: Embarrassingly parallel search. In:
  Schulte, C. (ed.) Principles and Practice of Constraint Programming. pp.
  596--610. Springer Berlin Heidelberg, Berlin, Heidelberg (2013)

\end{thebibliography}

\end{document}